\documentclass[11pt,a4paper]{article}
\usepackage[hyperref]{naaclhlt2018}
\usepackage{times}
\usepackage{latexsym}
\usepackage{wrapfig}
\usepackage{url}
\usepackage{graphicx}

\aclfinalcopy 
\def\aclpaperid{1127} 

\title{A Simple and Effective Approach to the Story Cloze Test}

\author{Siddarth Srinivasan, Richa Arora, Mark Riedl \\
 Georgia Institute of Technology \\
  {\tt \{sidsrini, richa.arora, riedl\}@gatech.edu}}

\date{}

\begin{document}
\maketitle
\begin{abstract}
  In the Story Cloze Test, a system is presented with a 4-sentence prompt to a story, and must determine which one of two potential endings is the `right' ending to the story. Previous work has shown that ignoring the training set and training a model on the validation set can achieve high accuracy on this task due to stylistic differences between the story endings in the training set and validation and test sets. Following this approach, we present a simpler fully-neural approach to the Story Cloze Test using skip-thought embeddings of the stories in a feed-forward network that achieves close to state-of-the-art performance on this task without any feature engineering. We also find that considering just the last sentence of the prompt instead of the whole prompt yields higher accuracy with our approach.
\end{abstract}

\section{Introduction}

\citet{mostafazadeh2016corpus} introduced the {\em Story Cloze Test}: given a four-sentence story prompt (or `context'), the task is to pick the `right' commonsense ending from two options. The Cloze Test is intended to be a general framework for evaluating story understanding, since it ostensibly requires combining semantic understanding and commonsense knowledge of our world. 
The task is accompanied by the Rochester story (ROCstory) corpus.
The training set consists of crowdsourced five-sentence stories designed to capture common events in daily life.
The validation and testing sets consist of four-sentence prompts and labeled `right' and `wrong' story endings. Table \ref{sample} shows such a sample story from the Rochester corpus validation set with a labeled right and wrong ending. 

Many previous approaches to the Cloze Test have ignored the training set entirely and trained on the validation set since the former lacks `negative' examples; although this greatly reduces the available training data, it circumvents the issue of obtaining negative examples during training.

Our contribution to this task is two-fold. 
First, we achieve near state-of-the-art performance (within $1.1\%$) but with a much simpler, fully-neural approach.
Where previous approaches rely on feature engineering or involved neural network architectures, we achieve high accuracy with a fully neural approach involving only a single feed-forward network and pre-trained skip-thought embeddings \citep{kiros2015skip}.
Second, we find that considering only the last sentence of the context outperforms models that consider the full context. Previous approaches focused on the accuracy achieved by either considering the whole context or ignoring the whole context of the story.
%
In sum, our approach differs from previous efforts in the joint use of three strategies: (1) using skip-thought embeddings \citep{kiros2015skip} for sentences in the story in a feed-forward neural network, (2) training the model on the provided validation set, and (3) considering the two endings with only the last sentence in the prompt. 

\begin{table}[t]
\small
\begin{center}
\begin{tabular}{|p{6 cm}|}
\hline
{\bf Story Context} \\
\hline
Bob loved to watch movies. \\
He was looking forward to a three day weekend coming up. \\
He made a list of his favorite movies and invited some friends over. \\
He spent the weekend with his friends watching all his favorite movies. \\
\hline
\emph{Right Ending:} Bob had a great time. \\
\emph{Wrong Ending:} Bob stopped talking to those friends. \\
\hline
\end{tabular}
\end{center}
\caption{\label{sample} A sample story from the ROCStory Validation Set }
\end{table}

This paper is structured as follows: we will discuss previous approaches to the problem and how they compare to our approach, describe our model and the experiments we ran in detail, and finally discuss reasons for our model's superior performance and why ignoring the first three sentences of the story produces better accuracy.

\section{Related Work}

\citet{mostafazadeh2016corpus} presented the original Story Cloze Test, and showed that while humans could achieve 100\% accuracy on the task, a deep structured semantic model \citep{huang2013learning} was the best performing artificial baseline, with a test-set accuracy of 58.5\%. While they do consider using skip-thought embeddings for this task, they do so by choosing the ending whose embedding was closer to the average skip-thought embedding of the context. This only achieves a test-set accuracy of 55.2\%. On the other hand, we train a feed-forward network using skip-thought embeddings.

The Story Cloze Test was the shared task at LSDSem 2017, and \citet{mostafazadeh2017lsdsem} summarize the approaches by various teams on this task. The best-performing system by \citet{schwartz2017story} achieved a test-set accuracy of 75.2\%. Like us, they train their model on the validation set, but their approach relies more heavily on feature engineering. 
They find that they could achieve 72.4\% accuracy using just the stylistic features of the endings, suggesting that many of the `right' endings on this task could be identified independent of the story context. Upon further investigation, \citet{Schwartz2017TheEO} find differences not only between the `right' and `wrong' endings in the validation set, but also between these and the `right' endings from the training set, providing some explanation for why models trained on the validation set outperform models trained on the training set - their data distributions are somewhat different.

Further work by \citet{Cai2017PayAT} established a neural baseline for models trained on the validation set, with a test-set accuracy of 74.7\%. They were also able to achieve a marginally better accuracy of 72.5\% (compared to \citet{schwartz2017story}) when using just the sentence endings and ignoring the context; and this approach did not require any feature engineering. They showed that a human can distinguish `right' from `wrong' endings \emph{without} the context with 78\% accuracy, further backing the claim that the importance of context in determining the right ending is more limited than desirable on this task. Their approach involves training a hierarchical bidirectional LSTM with attention to first encode sentences and then stories, with a hinge-loss objective function. 

\citet{roemmele2017rnn} use skip-thought embeddings for this task, but they encode the entire context using a GRU, with a binary classifier to determine if an ending was right or wrong. They train their model on the provided training set, sampling negative examples from the training set itself. Their best model achieves 67.2\% accuracy on this task.

Currently, the comprehensive approach taken by \citet{Chaturvedi2017StoryCF}, where they model event sequence, sentiment trajectory, and topical consistency for a hidden coherence model,  achieves the state-of-the-art performance on this task, with a test-set accuracy of 77.6\%.

\section{Approach}

We trained several models on both the training set and the validation set of the ROCStory corpus. When training a model on the training set, we obtain `negative' examples (i.e. wrong endings) by randomly choosing a sentence from another story in the corpus. In this section, we describe the choice of sentence embeddings, the architecture of the models we trained, and our experimental setup.

\subsection{Embeddings}

Key to our approach is the use of skip-thought embeddings \citep{kiros2015skip} in our feed-forward network (denoted {\bf skip} in Table \ref{results}). These are 4800-dimensional embeddings of sentences trained on the task of predicting their context using the BookCorpus dataset (a large dataset of books). We use a pre-trained skip-thought encoder\footnote{\texttt{https://github.com/ryankiros/skip-thoughts}} to obtain the embeddings for all sentences in the training set, validation set, and test set.

To isolate the increase in accuracy from using skip-thought vectors, we also experiment with learning sentence embeddings directly, for this task. Unlike the skip-thought encoder that directly gives sentence embeddings, we use a bidirectional LSTM that takes in GloVe embeddings \citep{pennington2014glove} of each word in the sentence and returns a 4800 dimensional embedding of the sentence (denoted {\bf GloVe} in Table \ref{results}) formed by concatenating the outputs of the forward and backward LSTMs. We use the GloVe model pre-trained on Wikipedia 2014 and Gigaword 5 data\footnote{\texttt{https://nlp.stanford.edu/projects/glove/}}.

\begin{figure*}
\begin{center}
\includegraphics[scale=0.35]{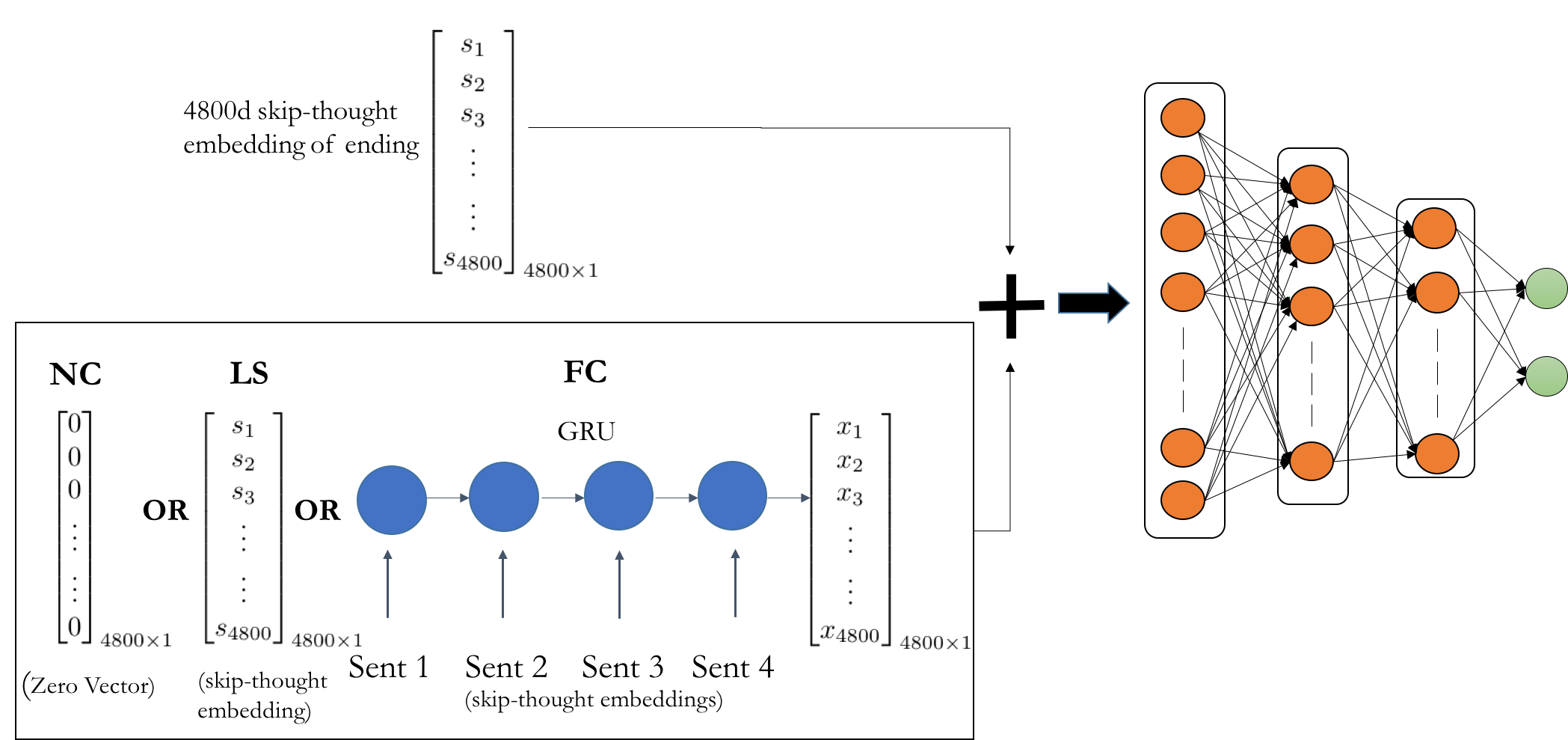}\caption{Model Architecture}
\end{center}
\end{figure*}

\subsection{Models}

Common to all our models is a single feed-forward neural network with a softmax-layer at the end that acts as a binary classifier. This neural network takes in a 4800-dimensional input (the same dimensionality as the skip-thought embeddings) and returns the probability of the endings being `right' and `wrong'. During inference time, we make a forward pass with each of the two possible endings, and select the ending that has a higher probability of being the `right' ending. 
We use two layer and three layer fully connected networks with Rectified Linear (ReLU) non-linearities (refer to Appendix A for model-specific architecture). We then experiment with different inputs to the neural network, as described below.

{\bf No Context (NC)} 
This model attempts to identify the `right' ending of a story by ignoring the story context and looking only at examples of right and wrong endings. As such, the input to the neural network is just the skip-thought embedding of the story ending, with 0/1 label indicating whether it was the `wrong' or `right' ending.

{\bf Last Sentence (LS)} 
In this model, the input to the neural network is the sum of the skip-thought embedding of the last sentence of the prompt (i.e., fourth sentence in the story) and the skip-thought embedding of the ending. Essentially, we are attempting to identify the right ending based on only the ending and the preceding sentence in the story.

{\bf Full Context (FC)} Here, we use a Gated Recurrent Unit (GRU) to encode the entire story prompt into a 4800-dimensional vector, add it to the skip-thought embedding of the story ending, and pass it as input to the neural network. The input to the GRU is the skip-thought embedding of each sentence, and this model attempts to identify the right ending by considering the entire story prompt.

\section{Experiments}
\label{sect:pdf}

\subsection{Dataset}

For all our experiments, we use the ROCStory corpus \cite{mostafazadeh2016corpus}. 
The corpus consists of a training set of 98,161 five-sentence stories, a validation set consisting of 1,871 four-sentence stories, and a test set of 1,871 four-sentence stories, with the validation and test sets providing labeled `right' and `wrong' story endings for each story.
\cite{mostafazadeh2016corpus} crowdsourced the collection of stories on Amazon Mechanical Turk; workers were asked to compose five-sentence stories about common daily situations with a clear beginning and end. 
To create the validation and testing sets, endings were removed from stories and an additional group of workers on Mechanical Turk were asked to provide a `right' ending or a `wrong' ending.

Although models trained on the validation set score higher than those trained on the training set as previously discussed, we provide the results for the same model trained on the validation set (denoted {\bf val}) as well as the training set (denoted {\bf trn}) in Table \ref{results} for comparison.

\subsection{Experimental Method}

When training on the training set, we tuned hyper-parameters using the validation set. When training on the validation set, we hold out 10\% of the validation set, and tune hyper-parameters to find a configuration that maximizes the accuracy on the held out data. We use cross-entropy loss and SGD with learning rate of 0.01. 

During training, we save the model every 3000 iterations, and calculate the validation accuracy. We train each model five times (except the FC models, which we train once due to time considerations), and report the average test set accuracy of the model. We use the model with the highest validation accuracy in each round to calculate the test set accuracy for that round. We present our results in Table \ref{results}.

\begin{table}[t!]
\begin{center}
\begin{tabular}{|c|c|c|c|}
\hline \bf model & \bf val & \bf test \\ \hline
trn-NC-skip & 60.3\% & 60.8\% \\
val-NC-skip & 73.9\% & 72.6\% \\ \hline
trn-FC-skip & 62.4\% & 62.6\% \\
val-FC-skip & 73.8\% & 71.6\% \\ \hline
trn-LS-skip & 62.8\% & 62.7\% \\
{\bf val-LS-skip} & {\bf 77.2\%} & {\bf 76.5\%} \\
val-LS-GloVe & 69.7\% & 63.0\% \\ \hline
\citet{Chaturvedi2017StoryCF} & - & \underline{77.6\%} \\
\citet{schwartz2017story} & - & 75.2\% \\
\citet{Cai2017PayAT} & - & 74.7\%\\
\hline
\end{tabular}
\end{center}
\caption{\label{results} Accuracies for various models on the Story Cloze Test}
\end{table}

\subsection{Results and Discussion}

The 3-layer feed-forward neural network trained on the validation set by summing the skip-thought embeddings of the last sentence (LS) of the story prompt and the ending gives the best accuracy ($76.5\%$). This approach is far simpler than previous approaches in the literature; it requires no feature engineering, nor intricate neural network architecture, and achieves close to state-of-the-art accuracy. Comparing `val-LS-skip' to `val-LS-GloVe' (i.e., using skip-thought embeddings for sentences vs. GloVe word embeddings), we confirm that the success of this approach lies in the sizable boost to accuracy from the use of pre-trained skip-thought embeddings.

This is perhaps unsurprising given the success of skip-thought embeddings in story-related tasks (\citet{Agrawal2016SortSS}, \citet{roemmele2017rnn}), since the model was trained on a large corpus of fiction. While the BookCorpus and ROCStories draw from different distributions, it is possible that skip-thought vectors implicitly encode a general notion of typical story continuation. In the absence of such a large dataset to learn such associations from, the LSTM with GloVe embedding inputs is unable to encode the necessary information to do well on this task.


We note that the model trained using only the last sentence (LS) of the story context has higher accuracy compared to the model that uses a GRU to encode the full context (FC), and even the \citet{Cai2017PayAT} model which encodes the entire context.
It is unclear from our experiments why this might be.
One hypothesis is that as stories near conclusion, the space of possible continuations contracts.
In the absence of further context, a default prior is assumed - as implicitly encoded in skip-thought vectors trained on BookCorpus - that is often correct. 
Providing more context may conflict with the default prior, introducing uncertainty.
Another hypothesis is that the Mechanical Turk workers creating the validation and test data sets focused more on the fourth sentence when writing their `right' and `wrong' endings, so once again, adding context introduces error.

Finally, we observe that the Story Cloze Test is an easier task than identifying whether a given ending is coherent or not, since the former involves a forced choice between two endings. During test time, the model does not need to classify whether a given ending is `right' or `wrong', as it learns to do during train time; instead, it simply needs to correctly predict which ending is \emph{less} wrong. 

\section{Conclusion}
\label{sec:length}
We have shown a simple yet effective neural model that achieves high accuracy on the Cloze Test, which is within $1.1\%$ of the state-of-the-art approach that relies on feature engineering. Additionally, we make a minor improvement on \citet{Cai2017PayAT}'s `ending-only' baseline accuracy of $72.5\%$ with our val-NC-skip model.
  
Finally, we also showed that, for the models tested here, using the full
context actually performs \emph{worse} than using just the last sentence of the context.
Future investigation will be needed to determine whether this is a property inherent to human storytelling or a form of bias introduced during data collection.

\vspace{-4mm}




\bibliography{naaclhlt2018}

\begin{thebibliography}{11}
\expandafter\ifx\csname natexlab\endcsname\relax\def\natexlab#1{#1}\fi

\bibitem[{Agrawal et~al.(2016)Agrawal, Chandrasekaran, Batra, Parikh, and
  Bansal}]{Agrawal2016SortSS}
Harsh Agrawal, Arjun Chandrasekaran, Dhruv Batra, Devi Parikh, and Mohit
  Bansal. 2016.
\newblock Sort story: Sorting jumbled images and captions into stories.
\newblock In \emph{EMNLP}.

\bibitem[{Cai et~al.(2017)Cai, Tu, and Gimpel}]{Cai2017PayAT}
Zheng Cai, Lifu Tu, and Kevin Gimpel. 2017.
\newblock Pay attention to the ending: Strong neural baselines for the roc
  story cloze task.
\newblock In \emph{ACL}.

\bibitem[{Chaturvedi et~al.(2017)Chaturvedi, Peng, and
  Roth}]{Chaturvedi2017StoryCF}
Snigdha Chaturvedi, Haoruo Peng, and Dan Roth. 2017.
\newblock Story comprehension for predicting what happens next.
\newblock In \emph{EMNLP}.

\bibitem[{Huang et~al.(2013)Huang, He, Gao, Deng, Acero, and
  Heck}]{huang2013learning}
Po-Sen Huang, Xiaodong He, Jianfeng Gao, Li~Deng, Alex Acero, and Larry Heck.
  2013.
\newblock Learning deep structured semantic models for web search using
  clickthrough data.
\newblock In \emph{Proceedings of the 22nd ACM international conference on
  Conference on information \& knowledge management}, pages 2333--2338. ACM.

\bibitem[{Kiros et~al.(2015)Kiros, Zhu, Salakhutdinov, Zemel, Urtasun,
  Torralba, and Fidler}]{kiros2015skip}
Ryan Kiros, Yukun Zhu, Ruslan~R Salakhutdinov, Richard Zemel, Raquel Urtasun,
  Antonio Torralba, and Sanja Fidler. 2015.
\newblock Skip-thought vectors.
\newblock In \emph{Advances in neural information processing systems}, pages
  3294--3302.

\bibitem[{Mostafazadeh et~al.(2016)Mostafazadeh, Chambers, He, Parikh, Batra,
  Vanderwende, Kohli, and Allen}]{mostafazadeh2016corpus}
Nasrin Mostafazadeh, Nathanael Chambers, Xiaodong He, Devi Parikh, Dhruv Batra,
  Lucy Vanderwende, Pushmeet Kohli, and James Allen. 2016.
\newblock A corpus and cloze evaluation for deeper understanding of commonsense
  stories.
\newblock In \emph{Proceedings of NAACL-HLT}, pages 839--849.

\bibitem[{Mostafazadeh et~al.(2017)Mostafazadeh, Roth, Louis, Chambers, and
  Allen}]{mostafazadeh2017lsdsem}
Nasrin Mostafazadeh, Michael Roth, Annie Louis, Nathanael Chambers, and James~F
  Allen. 2017.
\newblock Lsdsem 2017 shared task: The story cloze test.
\newblock \emph{LSDSem 2017}, pages 46--51.

\bibitem[{Pennington et~al.(2014)Pennington, Socher, and
  Manning}]{pennington2014glove}
Jeffrey Pennington, Richard Socher, and Christopher~D. Manning. 2014.
\newblock \href {http://www.aclweb.org/anthology/D14-1162} {Glove: Global
  vectors for word representation}.
\newblock In \emph{Empirical Methods in Natural Language Processing (EMNLP)},
  pages 1532--1543.

\bibitem[{Roemmele et~al.(2017)Roemmele, Kobayashi, Inoue, and
  Gordon}]{roemmele2017rnn}
Melissa Roemmele, Sosuke Kobayashi, Naoya Inoue, and Andrew~M Gordon. 2017.
\newblock An rnn-based binary classifier for the story cloze test.
\newblock \emph{LSDSem 2017}, pages 74--80.

\bibitem[{Schwartz et~al.(2017{\natexlab{a}})Schwartz, Sap, Konstas, Zilles,
  Choi, and Smith}]{Schwartz2017TheEO}
Roy Schwartz, Maarten Sap, Ioannis Konstas, Leila Zilles, Yejin Choi, and
  Noah~A. Smith. 2017{\natexlab{a}}.
\newblock The effect of different writing tasks on linguistic style: A case
  study of the roc story cloze task.
\newblock In \emph{CoNLL}.

\bibitem[{Schwartz et~al.(2017{\natexlab{b}})Schwartz, Sap, Konstas, Zilles,
  Choi, and Smith}]{schwartz2017story}
Roy Schwartz, Maarten Sap, Ioannis Konstas, Leila Zilles, Yejin Choi, and
  Noah~A Smith. 2017{\natexlab{b}}.
\newblock Story cloze task: Uw nlp system.
\newblock \emph{LSDSem 2017}, pages 52--55.

\end{thebibliography}
\bibliographystyle{acl_natbib}

\clearpage

\appendix

\section{Model Sizes}
Here we given more detail on the trained models from Table \ref{results}.

\begin{table}[h!]
\begin{center}
\begin{tabular}{|c|c|c|c|}
\hline \bf model & \bf \# HL & \bf Dim of HL & \bf Dim of enc\\ \hline
trn-NC-skip & $2$ & $256,64$ & - \\
val-NC-skip & $2$ & $256, 64$ & -\\ \hline
trn-FC-skip & $2$ & $256, 64$ & 4800\\
val-FC-skip & $2$ & $256, 64$ & 4800\\ \hline
trn-LS-skip & $3$ & $2400, 1200, 600$ & - \\
val-LS-skip & $3$ & $2400, 1200, 600$ & - \\
val-LS-GloVe & $3$ & $2400, 1200, 600$ & 4800 \\
\hline
\end{tabular}
\end{center}
\caption{\label{details} Description of Models trained (HL: Hidden Layer, enc: GRU or LSTM encoder used to encoder inputs)}
\end{table}

\end{document}